\documentclass[twocolumn,10pt]{article}

\usepackage{ifthen}
\newboolean{one_column}
\setboolean{one_column}{false}

\usepackage[utf8]{inputenc}
\usepackage[T1]{fontenc}
\usepackage{graphicx}
\usepackage{hyperref}
\usepackage{url}
\usepackage{breakurl}
\usepackage{xspace}
\usepackage{booktabs}
\usepackage[para,online,flushleft]{threeparttable}
\usepackage{multirow}
\usepackage{xcolor}

\usepackage{amsmath}
\usepackage{amssymb}

\usepackage{dsfont}

\usepackage[ucmark=true,hyperfirst=false]{glossaries}
\glsdisablehyper
\newacronym{1-D}{1-D}{1-Dimensional}
\newacronym{2-D}{2-D}{2-Dimensional}
\newacronym{ADC}{ADC}{Analog to Digital Converter}
\newacronym{AGC}{AGC}{Automatic Gain Control}
\newacronym{AI}{AI}{Artificial Intelligence}
\newacronym{AMHA}{AMHA}{\`A Mon Humble Avis}
\newacronym{ARAIM}{ARAIM}{Advanced Receiver Autonomous Integrity Monitoring System}
\newacronym{ASIC}{ASIC}{Application-Specific Integrated Circuit}
\newacronym{AWGN}{AWGN}{Additive White Gaussian Noise}
\newacronym{BAW}{BAW}{Bulk Acoustic Wave}
\newacronym{BOC}{BOC}{Binary Offset Carrier}
\newacronym{BPSK}{BPSK}{Binary Phase Shift Keying}
\newacronym{CNN}{CNN}{Convolutional Neural Network}
\newacronym{CNN-HL}{CNN-HL}{CNN-Histogram Loss}
\newacronym{CNN-Reg}{CNN-Reg}{CNN-Regression}
\newacronym{CPW}{CPW}{Coplanar Wave Line}
\newacronym{CW}{CW}{Carrier Wave}
\newacronym{DME}{DME}{Distance Measuring Equipment}
\newacronym{DC}{DC}{Direct frequency Conversion}
\newacronym{DC2}{DC}{Direct Current}
\newacronym{DF}{DF}{Dual-Frequency}
\newacronym{DL}{DL}{Deep Learning} 
\newacronym{DLL}{DLL}{Delay-Locked Loop}
\newacronym{DS}{DS}{Direct Sampling}
\newacronym{DSP}{DSP}{Digital Signal Processor}
\newacronym{ENAC}{ENAC}{\'Ecole Nationale de l'Aviation Civile}
\newacronym{ENC}{ENC}{European Navigation Conference}
\newacronym{FLL}{FLL}{Frequency Locked Loop}
\newacronym{FIR}{FIR}{Finite Impulse Response}
\newacronym{FPGA}{FPGA}{Field-Programmable Gate Array}
\newacronym{GBAS}{GBAS}{Ground-Based Augmentation System}
\newacronym{GLONASS}{GLONASS}{Global Navigation Satellite System}
\newacronym{GNSS}{GNSS}{Global Navigation Satellite System}
\newacronym{GPS}{GPS}{Global Positioning System}
\newacronym{GSL}{GSL}{GNU Scientific Library}
\newacronym{HL}{HL}{Histogram Loss}
\newacronym{ICD}{ICD}{Interface Control Document}
\newacronym{IF}{IF}{Intermediate Frequency}
\newacronym{iid}{i.i.d.}{Independent and Identically Distributed}
\newacronym{IIR}{IIR}{Infinite Impulse Response}
\newacronym{INS}{INS}{Inertial Navigation System}
\newacronym{ION}{ION}{Institute Of Navigation}
\newacronym{IRNSS}{IRNSS}{Indian Regional Navigational Satellite System}
\newacronym{KL}{KL}{Kullback-Leibler}
\newacronym{LEO}{LEO}{Low Earth Orbit}
\newacronym{LNA}{LNA}{Low-Noise Amplifier}
\newacronym{LOS}{LOS}{Line-Of-Sight}
\newacronym{LSTM}{LSTM}{Long Short-Term Memory}
\newacronym{MAE}{MAE}{Mean Absolute Error}
\newacronym{MEDLL}{MEDLL}{Multipath Estimating Delay Lock Loop}
\newacronym{ML}{ML}{Machine Learning}
\newacronym{MLP}{MLP}{Multi-Layer Perceptron}                                              
\newacronym{MOPS}{MOPS}{Minimum Operational Performance Specification}
\newacronym{MSB}{MSB}{Most Significant Bit}
\newacronym{NBI}{NBI}{Narrow Band Interference}
\newacronym{NCO}{NCO}{Numerically Controlled Oscillator}
\newacronym{NB}{NB}{Narrow Band}
\newacronym{NLOS}{NLOS}{Non Line-Of-Sight}
\newacronym{NN}{NN}{Neural Networks}
\newacronym{OCXO}{OCXO}{Oven Controlled Crystal Oscillator}
\newacronym{oktalse}{Oktal-SE}{Oktal-Synthetic Environment}
\newacronym{OS}{OS}{Open Service}
\newacronym{PCB}{PCB}{Printed Circuit Board}
\newacronym{PhD}{PhD}{Philosophi\ae Doctor}
\newacronym{PLL}{PLL}{Phase-Locked Loop}
\newacronym{PPS}{PPS}{Pulse Per Second}
\newacronym{PRN}{PRN}{Pseudo-Random Noise}
\newacronym{PSD}{PSD}{Power Spectral Density}
\newacronym{PVT}{PVT}{Position Time Velocity}
\newacronym{QPSK}{QPSK}{Quadrature Phase Shift Keying}
\newacronym{QZSS}{QZSS}{Quasi-Zenith Satellite System}
\newacronym{RF}{RF}{Radio Frequency}
\newacronym{RGB}{RGB}{Red Green Blue}
\newacronym{RHCP}{RHCP}{Right Hand Circular Polarization}
\newacronym{RMS}{RMS}{Root Mean Square}
\newacronym{RNN}{RNN}{Recurrent Neural Network}
\newacronym{RNSS}{RNSS}{Radio Navigation Satellite Service}
\newacronym{SAW}{SAW}{Surface Acoustic Wave}
\newacronym{SBAS}{SBAS}{Satellite-Based Augmentation System}
\newacronym{SDR}{SDR}{Software-Defined Radio}
\newacronym{SIS}{SIS}{Signal In Space}
\newacronym{SMA}{SMA}{SubMiniature version A}
\newacronym{SV}{SV}{Space Vehicle}
\newacronym{SVM}{SVM}{Support Vector Machines}
\newacronym{TCXO}{TCXO}{Temperature Controlled Crystal Oscillator}
\newacronym{THD}{THD}{Total Harmonic Distortion}
\newacronym{TIE}{TIE}{Time Interval Error}
\newacronym{VGA}{VGA}{Variable Gain Amplifier}
\newacronym{VGG}{VGG}{Visual Geometry Group}
\newacronym{VSWR}{VSWR}{Voltage Standing Wave Ratio}

\usepackage[thinqspace, squaren, Gray, cdot]{SIunits}

\usepackage{calc}
\usepackage{authblk}
\newlength{\mylength}
\ifthenelse{\boolean{one_column}}{
    \setlength{\mylength}{\textwidth/2}
    \usepackage{authblk}
}
{
    \setlength{\mylength}{\linewidth/\real{2.1}}
}

\DeclareGraphicsExtensions{.png,.eps,.pdf}

\newcommand{\dBHz}{\ensuremath{\deci\bel\hertz}\xspace}

\newcommand{\Hz}{\ensuremath{\hertz}\xspace}

\newcommand{\ms}{\ensuremath{\milli\second}\xspace}
\newcommand{\Tc}{\ensuremath{\textrm{Tc}}\xspace}

\newcommand{\CNZERO}{\ensuremath{\mathit{C\!/\!N0}}\xspace}

\newcommand{\ignore}[1]{}

\title{Distributional loss for convolutional neural network regression and application to GNSS multi-path estimation}

\author{Thomas Gonzàlez}
\affil{Oktal-Synthetic Environment, France \par ENAC, Université de Toulouse, France}
\author{Antoine Blais}
\affil{ENAC, Université de Toulouse, France} 
\author{Nicolas Couëllan}
\affil{ENAC, Université de Toulouse, France \par Institut de Mathématiques de Toulouse, France} 
\author{Christian Ruiz}
\affil{Oktal-Synthetic Environment, France}

\date{\today}

\begin{document}

\maketitle

\begin{abstract}
\gls{CNN} have been widely used in image classification. Over the years, they have also benefited from various enhancements and they are now considered as state of the art techniques for image like data. However, when they are used for regression to estimate some function value from images, fewer recommendations are available. In this study, a novel \gls{CNN} regression model is proposed. It combines convolutional neural layers to extract high level features representations from images with a soft labelling technique that helps generalization performance. More specifically, as the deep regression task is challenging, the idea is to account for some uncertainty in the targets that are seen as distributions around their mean. The estimations are carried out by the model in the form of distributions. Building from earlier work \cite{imani2018}, a specific histogram loss function based on the \gls{KL} divergence is applied during training. The model takes advantage of the \gls{CNN} feature representation and is able to carry out estimation from multi-channel input images. To assess and illustrate the technique, the model is applied to \gls{GNSS} multi-path estimation where multi-path signal parameters have to be estimated from correlator output images from the I and Q channels. The multi-path signal delay, magnitude, Doppler shift frequency and phase parameters are estimated from synthetically generated datasets of satellite signals. Experiments are conducted under various receiving conditions and various input images resolutions to test the estimation performances quality and robustness. The results show that the proposed soft labelling \gls{CNN} technique using distributional loss outperforms classical \gls{CNN} regression under all conditions. Furthermore, the extra learning performance achieved by the model allows the reduction of input image resolution from 80x80 down to 40x40 or sometimes 20x20.
\end{abstract}


\section{Introduction}
There has been a growing interest for \gls{CNN} \cite{deeplearningbook} in the machine learning community in order to construct high level features from images. Mostly used in images classification, \gls{CNN} have also been employed to estimate various information from images. Building regression models from image data is a difficult task since complex and high level feature representation is needed. This is why deep architectures are usually considered. Among the traditional examples from the literature, one can refer to \cite{Toshev2014DeepPoseHP}, \cite{Mahendran20173DPR}, \cite{Li2014DHP} and \cite{Girshick2011EfficientRO} where holistic reasoning on Human pose estimation is based on \gls{CNN}. Age estimation according to face images \cite{Yi2014AgeEB} or magnetic resonance images \cite{Ueda2019AnAE} are other successful examples of regression with \gls{CNN}. They were also applied on X-ray tensor images in \cite{Miao2016ACR} and recently, in \cite{lathuiliere2020}, the authors have proposed an extensive review of \gls{CNN} for regressions. 
To increase the regression model generalization performance, it has been shown that the use of soft labelling techniques can greatly help \cite{imani2018, gao2017}. Among soft labelling techniques, one idea is to consider that labels used during training are uncertain and drawn from a given distribution. Training can therefore be carried out at the distribution level rather than single observations. Especially suited when labels are ambiguous or subject to noise, this procedure can also be seen, in the general regression task, as a robustness enforcing alternative to other strategies such as batch normalization \cite{batchnorm}, dropout \cite{dropout}, early stopping \cite{earlystop} or regularization \cite{deeplearningbook, Couellan2021ProbabilisticRE, Couellan2021TheCE} most often used in classification problems.      

In this study, we propose a new \gls{CNN} regression model that implements distributional loss on \gls{GNSS} multi-path data. The objective is to estimate multi-path parameters from two dimensional \gls{GNSS} correlation images.

A multi-path is a parasitic reflection of the signal of interest which contaminates it at the very beginning of the receiving chain, the antenna. In the specific case of \gls{GNSS} receivers, multi-paths remain one of the most difficult disturbance to mitigate. Indeed, as the multi-path is of the same nature as the signal of interest it could be barely discernible from it. This similarity between the original signal and its disruptive replica can induce a large positioning error~\cite{Kos2010}. This problem explains the large number of research activities which have been led on multi-path detection, estimation and mitigation. Conventional signal processing methods have been extensively studied. In the statistical approach, the narrow correlator technique~\cite{VanDierendonck1992}, the early-late-slope technique~\cite{Townsend1994}, the strobe correlator~\cite{Garin1997}, the double-delta correlator~\cite{McGraw1999} and the multi-path insensitive delay lock loop~\cite{Jardak2011} methods have raised the interest of the \gls{GNSS} community, mainly for their simplicity despite their mixed efficiency. The Bayesian strategy was also explored, the \gls{MEDLL} remaining for years the reference implementation of the maximum likelihood principle~\cite{Townsend1995}, but with multiple variants proposed subsequently like in~\cite{Sahmoudi2008a} or \cite{BlancoDelgado2012}. However, the application of particle filtering to multi-path mitigation presented recently in~\cite{Qin2019} makes up for error accumulation of the \gls{MEDLL} algorithm, at the expense of a significantly higher computational complexity. It is worth noting that if the aforementioned methods are predominantly time-based, some research works have also investigated the frequency domain, through the Fourier transform~\cite{Zhang2004} or the wavelet decomposition~\cite{Zhang2004a}. Nevertheless, these methods may damage the signal of interest, particularly in cases where the spectrum of the multi-path is too close from the one of the direct path.
The use of \gls{ML} techniques to mitigate the errors in \gls{GNSS} signals has gained some interest in the early 2000s. A \gls{MLP} architecture designed to mitigate multi-path error for \gls{LEO} satellites has been detailed in~\cite{Vigneau2006} for example. More recently, taking advantage of the progress in kernel methods, ~\cite{Phan2013} proposed a support vector regressor using signal geometrical features to mitigate multi-path on ground fixed \gls{GPS} stations. Still with \gls{SVM}, \cite{Hsu2017} has conducted multi-path detection using high-level products of the GNSS receiver positioning unit. A comparison of the performances of \gls{SVM} and \gls{NN} algorithms to detect \gls{NLOS} multi-path is exposed in~\cite{Suzuki2021}, using native \gls{GNSS} signal processing outputs as features. Unsupervised \gls{ML} algorithms, like K-means clustering, have also been used with some success, for instance in~\cite{Savas2019}. However, the latest and significant advances in \gls{AI}, and notably in \gls{DL}, have opened up new perspectives. In~\cite{Quan2018}, using a \gls{CNN}, a carrier-phase multi-path detection model is developed. The authors propose to extract feature maps from multi-variable time series at the output of the signal processing stage using \gls{1-D} convolutional layers. \gls{DL} spoofing attack detection in \gls{GNSS} systems was addressed in the research literature~\cite{Schmidt2020} as well. Lately, \cite{Tao2021} has proposed a combined \gls{CNN}-\gls{LSTM} real-time approach, also based on \gls{1-D} convolution, and \cite{Kong2022} and \cite{Blais2022} have introduced the use of \gls{2-D} signal-as-image representation. \cite{Kong2022} makes use of \gls{MLP} input layers for automatic features construction whereas \cite{Blais2022} processes the images by \gls{2-D} convolutional filters. A review of the recent applications of \gls{ML} in \gls{GNSS} can also be found in~\cite{Siemuri2021}, focusing on use cases relevant to the \gls{GNSS} community.

However, the multi-path difficulty is still a challenge in \gls{GNSS} signal processing, especially concerning the most ambitious task, the multi-path removal. This ultimate solution requires an accurate estimation of the multi-path characteristics beforehand. To be more specific, the multi-path can be completely modeled by mean of four parameters, its delay, attenuation, frequency and phase. Therefore, this research work focuses on the estimation of these multi-path parameters through an original \gls{CNN} regression method.

The contributions provided in this work could be summarized as follows. To the best of our knowledge, this is the first model combining \gls{CNN} regression and distributional loss optimization that has been proposed. There is a growing literature on multi-path detection using \gls{ML} techniques, however, this study is one of the few research investigations, if not the first, that has been studying the application of modern \gls{DL} methods to \gls{GNSS} multi-path parameter estimation. It is also worth noticing that the regression model we propose works across input channels, using information from several distinct channels to carry out estimation. In that sense, the model is also different from models that learn from \gls{RGB} images that use replicas of the same input image under various color channels. Finally, this study shows clear evidence of boosted performance with the distributional regression loss model when compared to basic \gls{CNN} regression on the \gls{GNSS} multi-path application.

This article is organized as follows. Section \ref{CNN} details the concept of \gls{CNN} regression and distributional loss. Section \ref{sec:sec_gnss} describes the \gls{GNSS} multi-path parameter estimation problem, details the experiments that are conducted and provides numerical results. Section \ref{conclusions} concludes the article.

\section{\texorpdfstring{\gls{CNN}}{CNN} Regression}\label{CNN}
\subsection{Baseline \texorpdfstring{\gls{CNN}}{CNN}-Regression}

Regression methods~\cite{hastie} have been extensively studied and applied to many kinds of engineering problems. If data are not too complex, linear regression methods may be used. Alternatively, when direct linear models are not applicable, basis expansion may also be used in order to carry out linear regression on features instead of raw data. Mathematically, considering input vectors $x\in\mathcal{X}$, one would like to express a response $y\in\mathcal{Y}$ as a linear function of features $\phi_\lambda(x)\in\mathcal{H}$ as $y=w^\top\phi_\lambda(x)$ where $w$ are the regression weights and $\lambda$ represents the parameters needed to represent $x$ by its features in $\mathcal{H}$. There are several techniques to construct the features $\phi_\lambda(x)$ such as basis function expansion to carry out polynomial regression, methods for \gls{SVM} \cite{murphy} - in theses two cases $\lambda$ may represent degrees of polynomials or parameters of kernel functions respectively - or also \gls{NN} representations. 

In this study, we are interested in the specific case where $\mathcal{X}$ is a subspace of images that can be represented in $\mathbb{R}^{r1\times r2\times n_C}$ where $r1\times r2$ is the resolution of the images and $n_C$ is the number of channels (ex: $n_C=3$ for \gls{RGB} images or $n_C=2$ for \gls{GNSS} correlation images on I and Q channels in our study). For such input objects, it has been shown empirically that the construction of features $\phi_\lambda(x)$ using convolutional filters achieves the best feature representation for image-like inputs~\cite{lecun}. This is the reason why several authors have proposed convolutional network regression techniques to estimate various information from images (see~\cite{lathuiliere2020} for a review of such techniques). 

\begin{figure}
	\includegraphics[width=\linewidth]{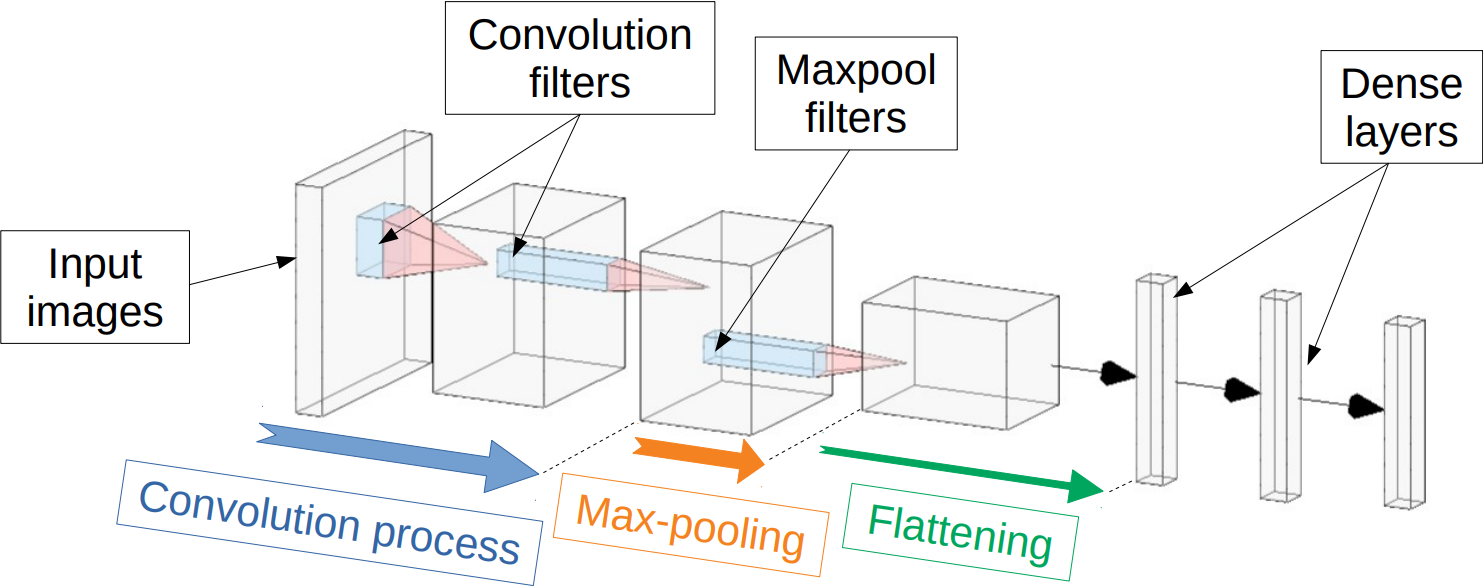}
	\caption{Image convolutional processing inside a \gls{CNN}.
	Image generated with~\cite{nnsvg}.}
	\label{fig:fig_VGG}
\end{figure}

\gls{CNN} are composed of several layers. Multiple convolutional blocks are used in order to extract local features in the images using a variety of parameterized convolutional filters (see~\cite{deeplearningbook} for a detailed explanation of convolutional mechanisms). Successive convolutional layers combined with activation layers (usually rectified unit activation layers) capture multiscale features in the images. Stacking several convolutional blocks that are separated by max-pooling layers to reduce dimension will allow the extraction of highly complex features $\phi_\lambda(x)$ where $\lambda=(W^l,b^l)_{l=1,\ldots,L}$, $W^l$ are the filter weights tensors and $b^l$ are the bias vectors for the $L$ convolutional layers. These features are then fed to conventional dense layers that are fully connected with connection weights and compute the regression model $y=w^\top\phi_\lambda(x)$. There are various architectures that can be designed for such purposes. Among these, \gls{VGG}-like networks~\cite{VGG} are often used since they have proven to be very effective in practice. The idea is to increase the number of convolutional filters when going deeper in the architecture. See Figure~\ref{fig:fig_VGG} for an example of a basic \gls{CNN} architecture.\\

The \gls{CNN} regression task on complex input data such as images is intrinsically difficult. The training of deep representations of features combined with the training of a dense network on a large flatten feature representation are required. Sometimes, the image information is also spread across several input channels as we will see later when dealing with I and Q \gls{GNSS} correlation images (see section \ref{sec:sec_gnss}).
Therefore, the regression model has to construct complex features representations from the various channels, adding even more complexity to the overall regression process. In order to ease the training of such multi-channel regression, in the next section, we will propose to use a soft label procedure. The main idea is to learn soft labels rather than sharp target continuous values.

Note that in the following, the response of the complete neural network regression model will be written as $y=N_\theta(x)$ where $\theta=(\lambda,W)$ and $\lambda$ represent filter and connection weights from the \gls{CNN} network architecture up to the penultimate dense layer and $W$ is the weight matrix storing all connection weights from the last dense layers.\\

\begin{figure*}
    \centering
    \includegraphics[width=\linewidth]{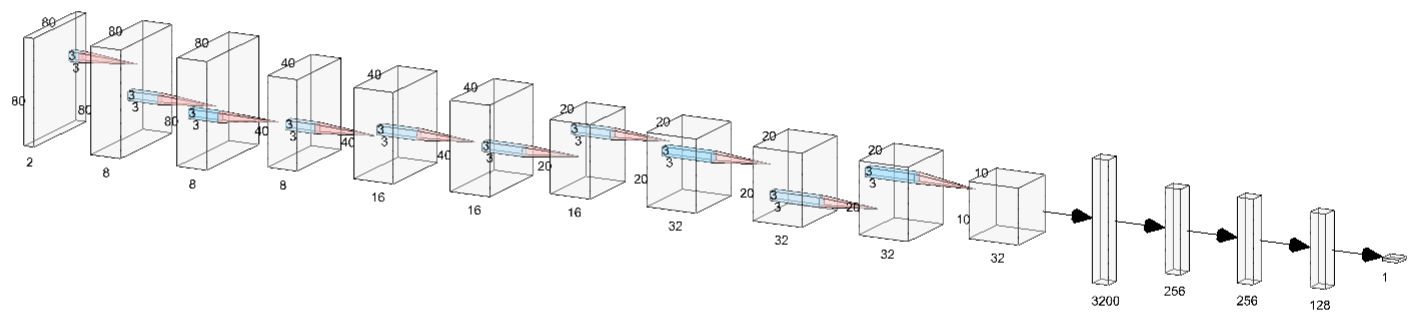}
    \caption{\gls{CNN} architecture used for regression task on 80x80 images. Image generated on \cite{nnsvg}.}
    \label{fig:fig_CNNReg}
\end{figure*}

\subsection{Soft labelling using distributional loss}\label{sub_sec:sub_sec_softlab}
In this section, we recall the concept of distributional loss as proposed in \cite{imani2018}. Traditionally, given an input $x$, the regression task consists in computing the parameter $\theta$ of a regression model $N_\theta$ that would assign predicted values $N_\theta(x)$ for given inputs $x$ and targets $y$. The target values $y$ can be seen as expected values of an underlying distribution of $Y|x$ assumed to be Gaussian. Therefore, in this setting, it is natural to calibrate the parameter $\theta$ by minimizing a square loss function $(N_\theta(x)-y)^2$.\\
The target value $y$ is usually considered as the ground truth and obtained by experiments or measurements. However, it may be subject to uncertainty or ambiguity~\cite{gao2017} and may impair the generalization performance of the regression model. The main idea of soft labelling is to consider that the target value is an observation of an underlying ground truth distribution and one may benefit from learning the distribution rather than the individual targets.\\
To do so, consider now the task of learning the distribution $Y|x$ instead of predicting directly $\mathbb{E}[Y|x]$. For each target value, a target distribution $Y|x$ is chosen. This choice is of course problem dependent. However in~\cite{imani2018}, it has been shown experimentally, when no prior knowledge on the target distribution is known, that the use of a truncated Gaussian is the best choice among a variety of other distributions. In the sequel of the article, we will therefore assume that $Y|x\sim \mathcal{N}_{[a,b]}(\mu,\sigma^2)$ where $\mathcal{N}_{[a,b]}(\mu,\sigma^2)$ is the truncated Gaussian distribution of mean $\mu$ and variance $\sigma^2$ on the interval $[a,b]$ with density
$$
f_{\mu,\sigma,[a,b]}(y)=\frac{1}{\sigma}\frac{\varphi(\frac{y-\mu}{\sigma})}{\phi(\frac{b-\mu}{\sigma})-\phi(\frac{a-\mu}{\sigma})}\mathds{1}_{[a,b]}(y)
$$
where $\varphi$ and $\phi$ are the density and the cumulative distribution functions of the standard normal distribution $\mathcal{N}(0,1)$.\\
In order to learn the target distributions, a discrete version of $Y|x$ is considered in the form of an histogram with $K$ bins.
During training, as we are now comparing the estimated distribution $N_\theta(x)$ and the target distribution $f_{\mu,\sigma,[a,b]}$, we can use the \gls{KL} divergence as a loss function. It is defined as follows
$$\mathcal{D}_{\mathit{KL}}(p||q)=\int_{-\infty}^{+\infty}p(y)\log\frac{p(y)}{q(y)}dy $$ and measures how different $q$ is from $p$. Using $\mathcal{D}_{\mathit{KL}}$ as a loss function will require to minimize simply the following term also known as the cross-entropy
$$
h(p,q)=-\int_{-\infty}^{+\infty}p(y)\log q(y)dy.
$$
If $q$ is the estimated density and $p$ is the target distribution with density $f_{\mu,\sigma,[a,b]}$, the cross-entropy can be written as
$$
h(f_{\mu,\sigma,[a,b]},q)=-\int_{a}^{b} f_{\mu,\sigma,[a,b]}(y)\log q(y) dy.
$$
For each input $x$, the estimated discrete distribution $q$ is a $K$-dimensional vector where each component $i$ returns a probability $q_i=(N_\theta(x))_i$ that the target value is in the bin $i$. The number of bins $K$ will be an hyper-parameter of the method. The discrete target distribution $p$ corresponding to $f_{\mu,\sigma,[a,b]}$ can also be seen as a $K$-dimensional vector where each component $i$ returns the probability mass $p_i=F(a_i+t_i)-F(a_i)$ contained in the $i$-th bin, where $a_i$ and $t_i$ are the left bound and the width of the bin respectively and $F$ is the cumulative distribution of $f_{\mu,\sigma,[a,b]}$. Therefore, for a given input $x$ the distributional \gls{HL} can be written as follows
$$
\mathit{HL}(x)=-\displaystyle\sum_{i=1}^K p_i \log (N_\theta(x))_i
$$
$$
\mathit{HL}(x)=-\sum_{i=1}^K \left[F(a_i+t_i)-F(a_i)\right] \log (N_\theta(x))_i
$$
After training, for a given input $x$, $\mathbb{E}[N_\theta(x)]$ will provide the estimated value of $\mathbb{E}[Y|x]$. In~\cite{imani2018, gao2017}, experiments have shown that learning target distributions rather than hard target values will improve the generalization power of the model. This method is different from transforming the regression task into a classification task by discretizing uniformly the support set of target values. Here, a target distribution is assumed and learned during the process. Additionally, by tuning the variance parameter $\sigma$ and the number of bins used to transform the target distribution into an histogram, it is possible to adjust the bias-variance trade-off achieved by the model. 

\subsection{Dedicated \texorpdfstring{\gls{CNN}}{CNN} architectures}
In this study, in order to compare the \gls{HL} strategy with the classical \gls{CNN-Reg}, two \gls{VGG}-like architectures~\cite{VGG} will be used. Sharing the same backbone as shown in Figure~\ref{fig:fig_CNNReg}, they only differ in their outputs. The extraction part is composed by 3 convolution blocks, each formed by consecutive convolutional layers and ended by a max-pooling layer. The number of filters is increasing with the depth of the convolutional layers. At the end of the convolutional process, the data are flattened to feed the 3 hidden dense layers. The output layer consists in a single neuron equipped with a linear activation function for the baseline \gls{CNN-Reg}, and a softmax layer with $K$ outputs in the soft labelling case. To differentiate the two algorithms, we will refer to the network with a single output as \gls{CNN-Reg} and the one using the softmax output as \gls{CNN-HL}.

\section{Application to \texorpdfstring{\gls{GNSS}}{GNSS} multi-path parameter estimation}\label{sec:sec_gnss}
\subsection{Problem statement}\label{sec:sec_problem_statement}

The \gls{GNSS} positioning principle is based on the distance measurement between satellites of known positions and the receiver to locate. Using the propagation time of a dedicated signal emitted by the satellite, the receiver estimates its relative distance to the satellite~\cite{Tsui2005}. Using several distances, the receiver is able to calculate its position by trilateration.

More precisely, the calculation process of the \gls{PVT} solution relies on the synchronisation between the received signal and a receiver replica of the wanted signal. The alignment between both signals requires the estimation of the three unknown parameters of the incoming signal which are:
\begin{itemize}
    \item The propagation delay $\tau$,
    \item The Doppler shift frequency $f_{\mathit{D}}$,
    \item The carrier phase $\phi$.
\end{itemize}
In a classical receiver, this estimation process is typically conducted by mean of the maximum likelihood principle. It is implemented through the maximization of the cross-correlation between the received signal and a local replica signal parameterized by three test values $\tilde{\tau}$, $\tilde{f_{D}}$ and $\tilde{\phi}$. In practice, the correlation operation is accomplished through a product followed by an integrate and dump stage. To be complete, it should also be pointed out that correlation is split into two orthogonal channels, named by convention In-phase (I) and in-Quadrature (Q). The maximization task is generally carried out by loop systems, namely the \gls{DLL} and the \gls{PLL}. In addition, the loops present the intrinsic advantage to track the target parameters which are time-varying due to the continuously changing receiver-satellites geometry. No more than three correlation points per I and Q channels are usually required for normal loop operation.

\begin{figure}
	\centering
	\includegraphics[width=\linewidth]{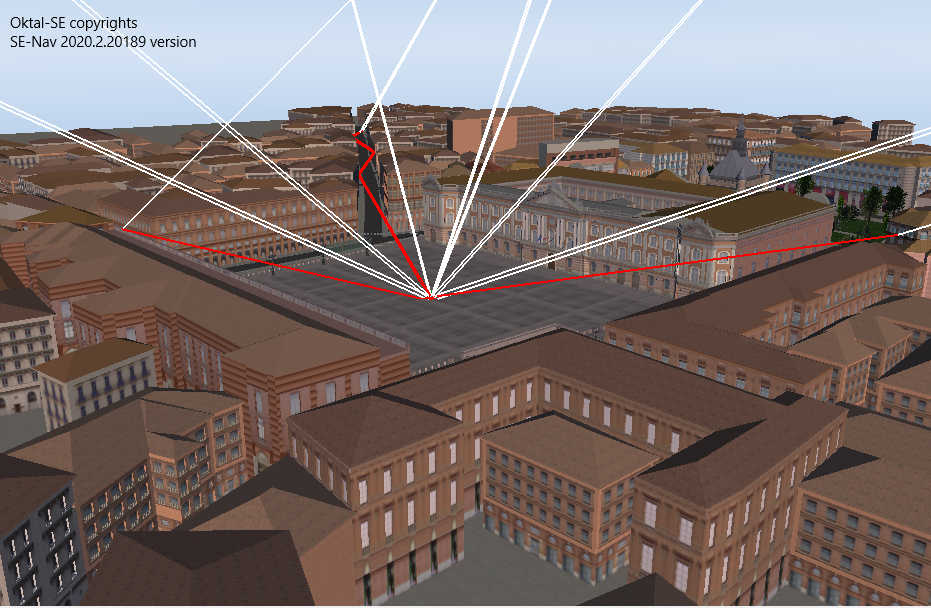}
	\caption{\gls{GNSS} receiver collecting \gls{LOS} and \gls{NLOS} signals in Toulouse city center (France).}
	\label{fig:fig_capitole}
\end{figure}

However, various kinds of interference can deteriorate the positioning process. Multi-path is one of the most common and the most harmful interference. A multi-path is a reflection on a surrounding obstacle of the useful signal picked up by the receiving antenna concurrently to the \gls{LOS} signal. An illustration of the phenomenon is shown in Figure~\ref{fig:fig_capitole}, generated by the SE-Nav software~\cite{senav}, a signal propagation simulator. In general, a receiver is impacted by multiple multi-paths, especially in urban environments where reflectors are numerous. Sometimes, the direct path may even be absent due to an obstruction, for example when high buildings are surrounding the receiver~\cite{Ziedan2018}. However, in this study the assumption is made that the direct path is always present and a single multi-path will be considered. Being the replica of the signal of interest, the multi-path contains the same information but with shifted parameters:
\begin{itemize}
    \item The code delay in excess compared with the useful signal $\Delta \tau_{\mathit{MP}}$,
    \item The difference in Doppler shift frequency with the useful signal $\Delta f_{\mathit{MP}}$,
    \item The phase difference with the useful signal $\Delta \phi_{\mathit{MP}}$,
    \item The attenuation of the multi-path with respect to the magnitude of the useful signal $\alpha_{\mathit{MP}}$.
\end{itemize}

The addition of the multi-path to the direct signal biases the result of the correlation operation. As a consequence, the estimation of the incoming signal parameters is altered and the accuracy of the position delivered to the user may be degraded. With as few as three correlation points per channel, the information available to detect the multi-path contamination and possibly mitigate its effects is poor.

\begin{figure}
	\centering
	\includegraphics[width=\linewidth]{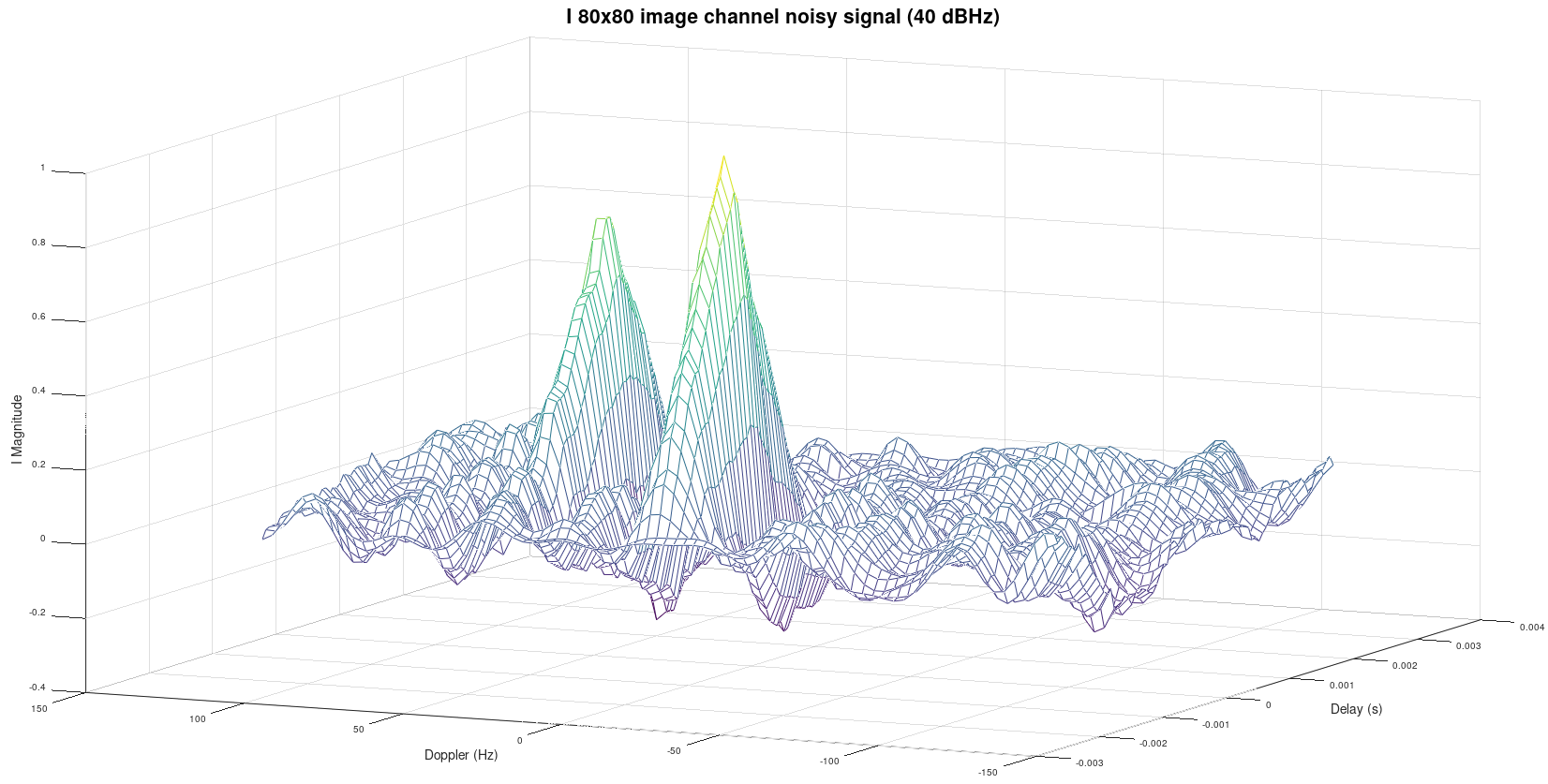}
	\caption[Caption for LOF]{I channel for a signal undergoing a multi-path with $\tau_{\mathit{MP}} = 0.75$ \Tc\footnotemark[1], $f_{\mathit{MP}} = 100$ \Hz, $\alpha_{\mathit{MP}} = 0.8$, $\phi_{\mathit{MP}} = 45^\circ$.}
	\label{fig:fig_I_channel}
\end{figure}

\begin{figure}
	\centering
	\includegraphics[width=\linewidth]{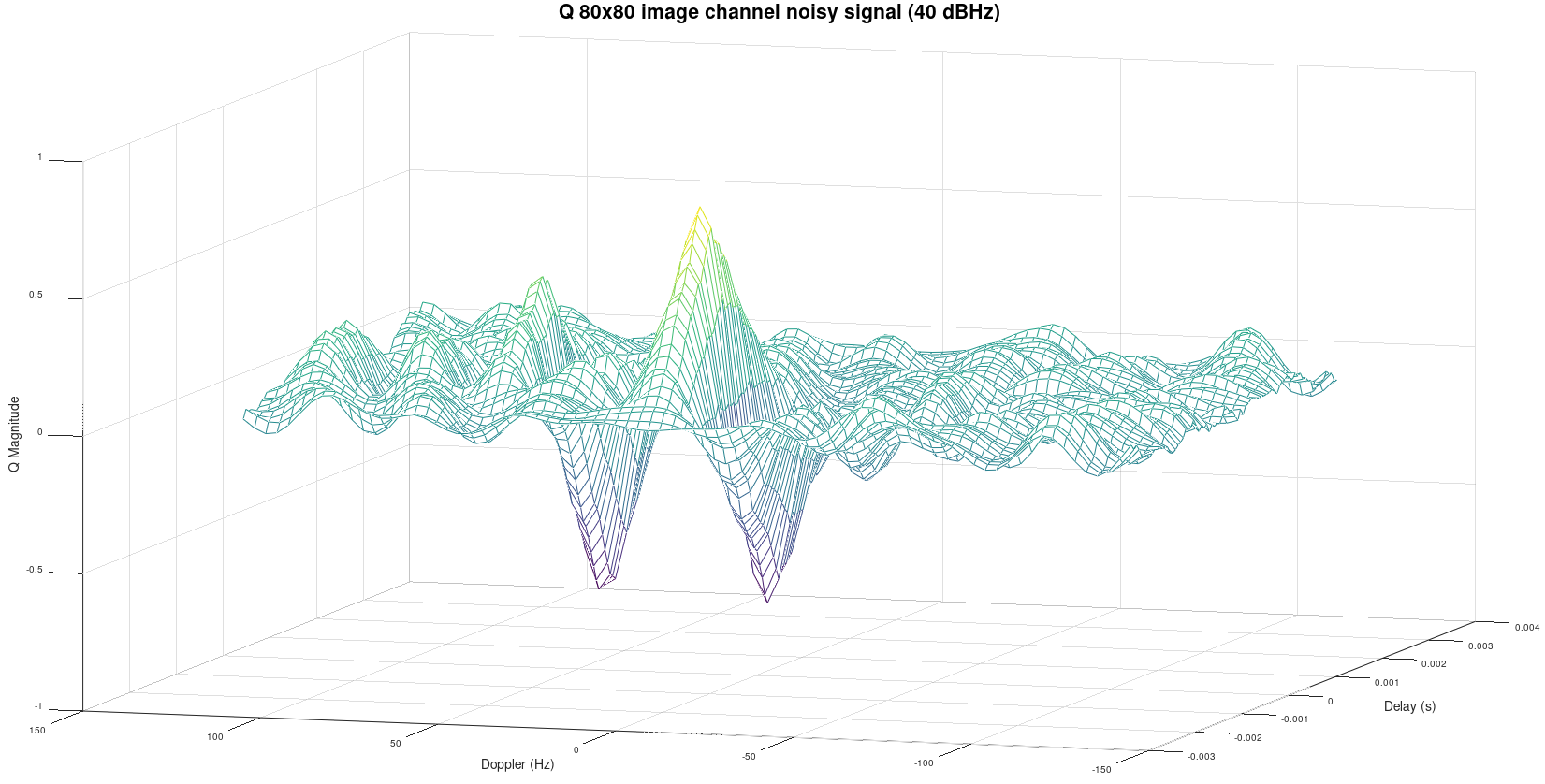}
	\caption{Q channel for a signal undergoing a multi-path with $\tau_{\mathit{MP}} = 0.75$ \Tc, $f_{\mathit{MP}} = 100$ \Hz, $\alpha_{\mathit{MP}} = 0.8$, $\phi_{\mathit{MP}} = 45^\circ$.}
	\label{fig:fig_Q_channel}
\end{figure}

\begin{figure}
\includegraphics[width=\linewidth]{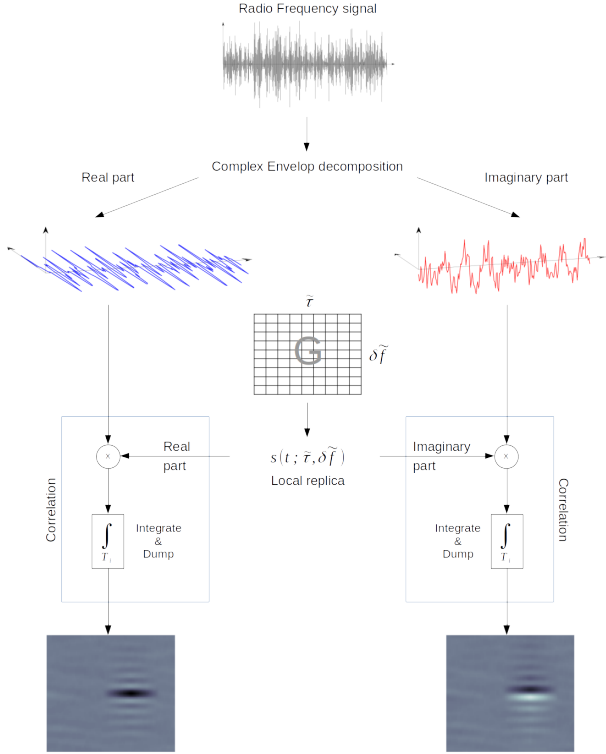}
\caption{Illustration of the image elaboration process. The received signal is split in two orthogonal components. Next, each component is correlated with a local replica signal whose parameters span a grid G. The result forms a 2D-image. The pair of images then supplies the \gls{CNN} implementing the regression task. The tilde notation indicates the local parameter by opposition to the received signal unknown parameter.}
\label{fig:fig_r_I_Q}
\end{figure}

This work proposes to use a larger number of correlation points in order to overcome this lack of information. Indeed, $\tilde{\tau}$ and $\tilde{f_{\mathit{D}}}$ each sample a specific range, forming a \gls{2-D} grid. The correlation outputs in turn compose a \gls{2-D} matrix, in other words an image. The signal having 2 channels, I and Q, there will be two 2D-images at the output of the correlators. The general process is depicted in Figure~\ref{fig:fig_r_I_Q}. Figures~\ref{fig:fig_I_channel} and \ref{fig:fig_Q_channel} are given as detailed examples of the resulting images. Concerning the estimation of $\phi$, the necessary information is available through the orthogonality property of the I and Q channels, the first one granting access to $\cos(\phi)$ and the second to $\sin(\phi)$. The image construction is detailed in \cite{Blais2022}.

\footnotetext[1]{The results presented in this paper where established using the \gls{GPS} L1 C/A legacy signal. However, the authors are confident that they could be generalized to other navigation signals, with the same overall structure, as no specific assumption has been made. $\Tc$ is the chip period, a basic defining parameter of this type of signal. $\Tc = 1/1023$ \ms for the \gls{GPS} L1 C/A signal.}

\subsection{Experimental setup}

\subsubsection{Dataset definition}
Experiments were made on several datasets which are divided into 2 groups. The first group is characterized by a fixed Carrier to Noise \CNZERO ratio while the second has distinct \CNZERO ratio levels. The \CNZERO figure represents the ratio between the power of the signal of interest and the receiver intrinsic noise level. In the \gls{GNSS} community it is considered as one of the most important figure of merit for the quality of the received signal. The higher the ratio, the more accurate the downstream parameter estimation.

In the first group, \CNZERO is equal to 40 \dBHz to simulate a typical urban environment receiving condition. The multi-path attenuation $\alpha_{\mathit{MP}}$ is categorized to assess the performance of both \gls{CNN} algorithms as a function of this parameter.
In this way, we have:
\begin{itemize}
    \item \textit{Strong multi-paths} where $\alpha_{\mathit{MP}} \sim \mathcal{U}([0.6, 0.9])$
    \item \textit{Moderated multi-paths} where $\alpha_{\mathit{MP}} \sim \mathcal{U}([0.4, 0.6])$
    \item \textit{Weak multi-paths} where $\alpha_{\mathit{MP}} \sim \mathcal{U}([0.1, 0.4])$
\end{itemize}

The second group is composed by datasets with different \CNZERO ratio while $\alpha_{\mathit{MP}} \sim \mathcal{U}([0.1, 0.9])$. The goal is to evaluate the algorithms with respect to \CNZERO, which takes the following values: 43 \dBHz, 40 \dBHz, 37 \dBHz and 34 \dBHz.

Otherwise, all datasets share the same following variables:
\begin{itemize}
    \item $\Delta\tau_{\mathit{MP}}$ is uniformly generated in $[0, 1.5]$ \Tc,
    \item $\Delta f_{\mathit{MP}}$ follows a truncated centered Gaussian distribution with standard deviation $\sigma=125/3$ \Hz. $\Delta f_{\mathit{MP}}$ is restricted to $[-3\sigma, +3\sigma]$ \Hz,
    \item $\Delta\phi_{\mathit{MP}}$ is uniformly generated in $[0, 2\pi]$ \rad.
\end{itemize}

Each dataset contains $10000$ samples, a sample being the I and Q image pair. In a dataset, image size does not change. However, to observe the \gls{CNN} parameter estimation performance under lower input image resolutions, various datasets were created: 80x80, 40x40, 20x20 and 10x10 pixels images.

A synthetic correlator output generator~\cite{Blais2021} has been used to populate the datasets used in this study. The underlying signal model and the generation process are completely described in~\cite{Blais2022}. The generator is fully configurable with respect to the multi-path parameters \CNZERO, $\alpha_{\mathit{MP}}$, $\Delta\tau_{\mathit{MP}}$, $\Delta f_{\mathit{MP}}$ and $\Delta\phi_{\mathit{MP}}$. The resolution of the I \& Q output images can also be set on demand.

\subsubsection{Hyper-parameter selection}
Performance results provided in the next section are results averaged over 10 runs. For each run, the training set is formed by randomly selecting $8000$ samples and the validation set by selecting randomly $1000$ samples among the $2000$ remaining samples. The last $1000$ samples constitute the test set.

\gls{CNN-Reg} and \gls{CNN-HL} were trained for 100 epochs and have used a $1000$ samples batch size. Learning rates of both \gls{CNN} were empirically fixed to $10^{-3}$ and 3x3 filters were used for convolutional layers as shown in Figure~\ref{fig:fig_CNNReg}. The number of bins for \gls{CNN-HL} is fixed to $K=100$ (such as in~\cite{imani2018}). The hyper-parameter $\sigma$ follows the following rule: $\sigma=2(b-a)/K$.

\subsection{Results}

Performance assessment is based on \gls{MAE} results averaged over 10 runs. For the \gls{CNN-Reg}, the \gls{MAE} metric is as follows: 
$$
\frac{1}{L}\sum_{l=1}^{L} | N_\theta(x_l) - y_l|
$$
where:
\begin{itemize}
    \item $L$ is the number of samples in the test dataset,
    \item $x_l$ is the $l$-th test sample,
    \item $y_l$ is the target value for the $l$-th sample.
\end{itemize}
For the \gls{CNN-HL}, the \gls{MAE} metric is calculated as follows:
$$
\frac{1}{L}\sum_{l=1}^L \left|\tilde{\mathbb{E}}[N_\theta(x)] - y_l\right| 
$$
with
$$
\tilde{\mathbb{E}}[N_\theta(x)] = \sum_{i=1}^K q_{i,x_l}.c_i
$$
and
\begin{itemize}
    \item $q_{i,x_l}$ is the probability of the $i$-th softmax output neuron associated to $x_l$,
    \item $c_i$ is the center of the $i$-th bin for $i \in [1, K]$ (where the interval $[a,b]$ has been partitioned into $K$ equal subdivisions).
\end{itemize}

Tables~\ref{tab:delay_table}, \ref{tab:doppler_table}, \ref{tab:alpha_table} and \ref{tab:phase_table} gather the averaged \gls{MAE} performance results as defined above for the two algorithms \gls{CNN-Reg} and \gls{CNN-HL} on the first group of datasets.

\begin{figure}
    \centering
    \begin{tabular}{cc}
    \includegraphics[width=\mylength]{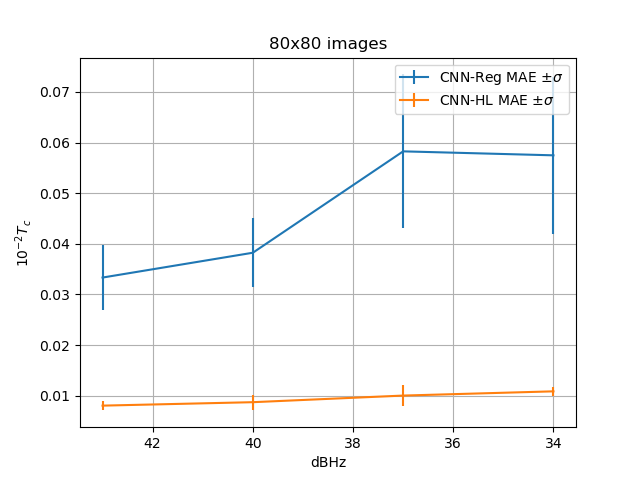} & 
    \includegraphics[width=\mylength]{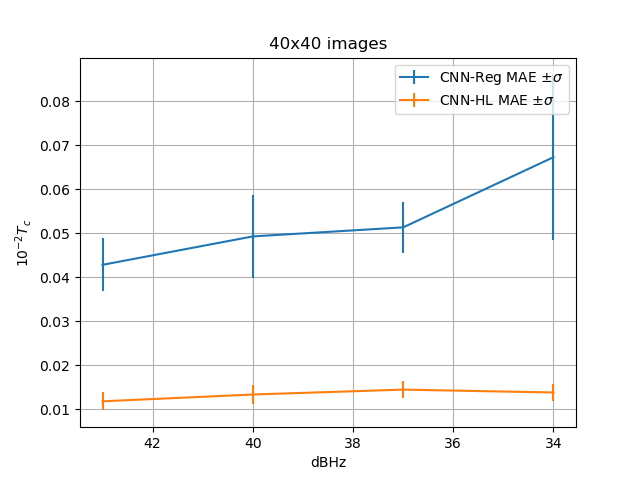} \\
    \includegraphics[width=\mylength]{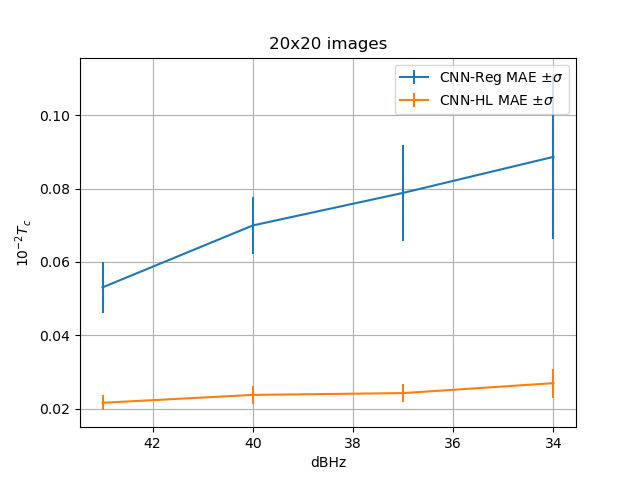} &
    \includegraphics[width=\mylength]{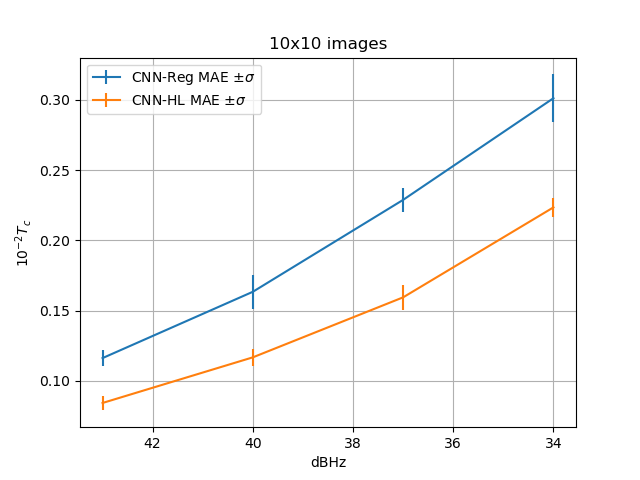} \\
    \end{tabular}
    \centering
    \caption{\gls{MAE} behaviour according to \CNZERO ratio for $\tau_{\mathit{MP}}$ estimation.}
	\label{fig:fig_delay}
\end{figure}

\begin{figure}
    \centering
    \begin{tabular}{cc}
    \includegraphics[width=\mylength]{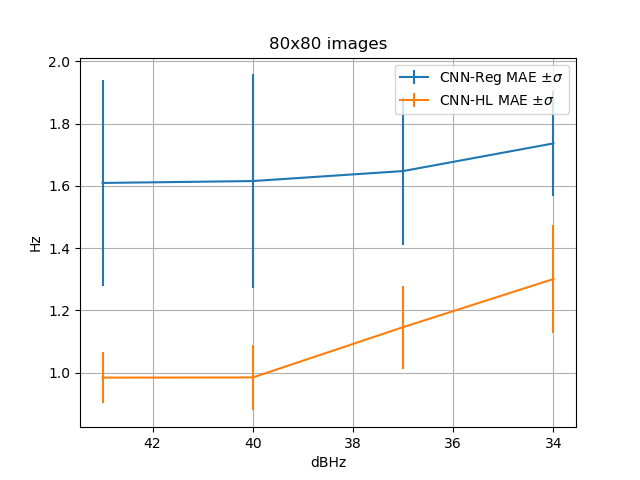} & \includegraphics[width=\mylength]{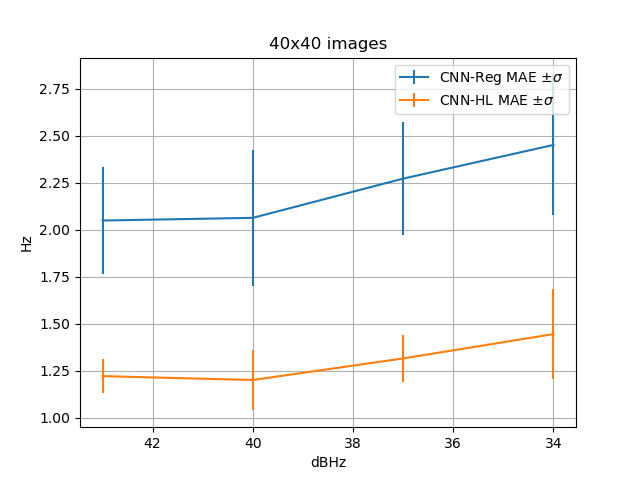} \\
    \includegraphics[width=\mylength]{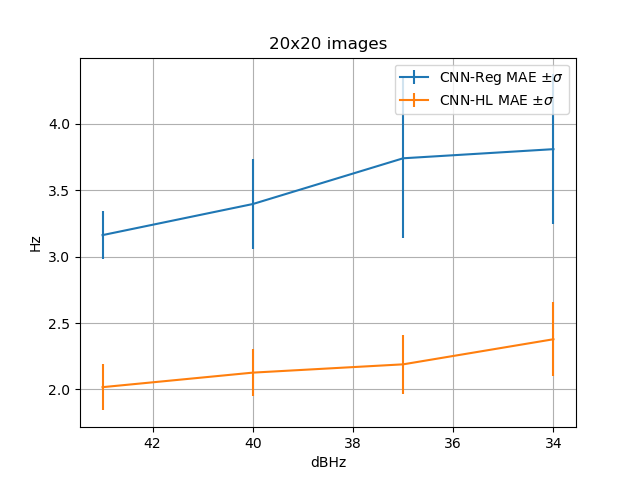} &
    \includegraphics[width=\mylength]{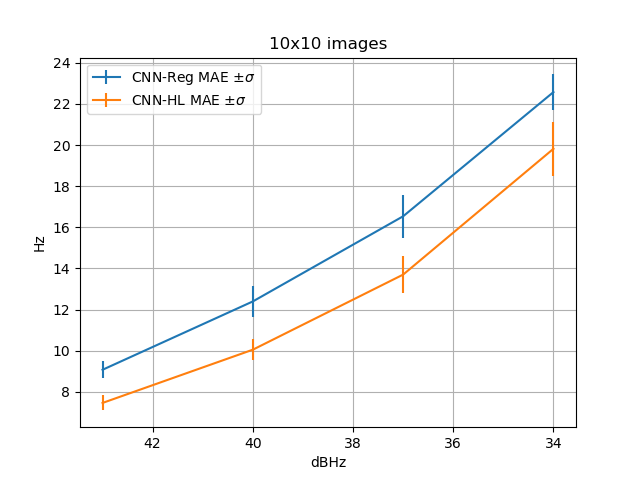} \\
    \end{tabular}
    \centering
    \caption{MAE behaviour according to \CNZERO ratio for $f_{\mathit{MP}}$ estimation.}
	\label{fig:fig_dopp}
\end{figure}

\begin{figure}
    \centering
    \begin{tabular}{cc}
    \includegraphics[width=\mylength]{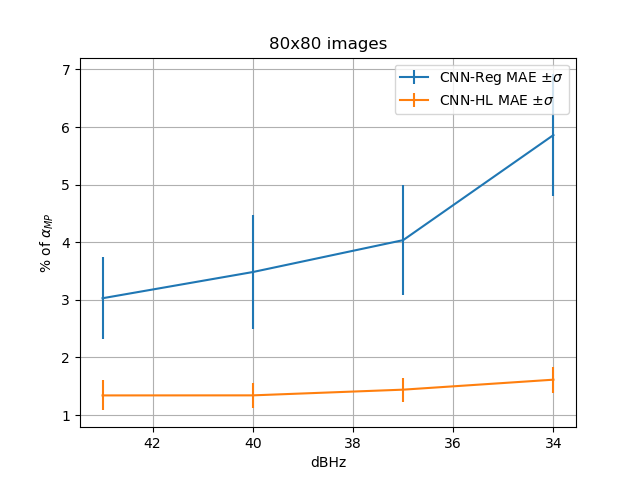} & \includegraphics[width=\mylength]{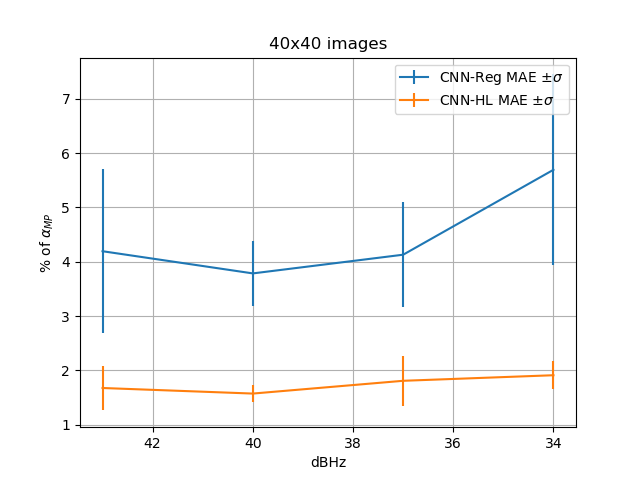} \\
    \includegraphics[width=\mylength]{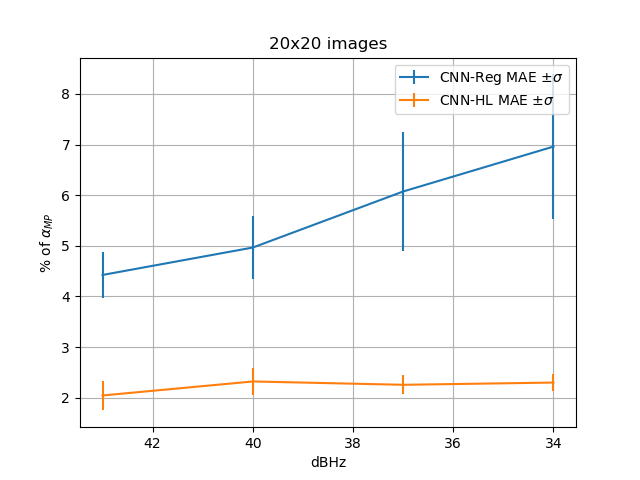} &
    \includegraphics[width=\mylength]{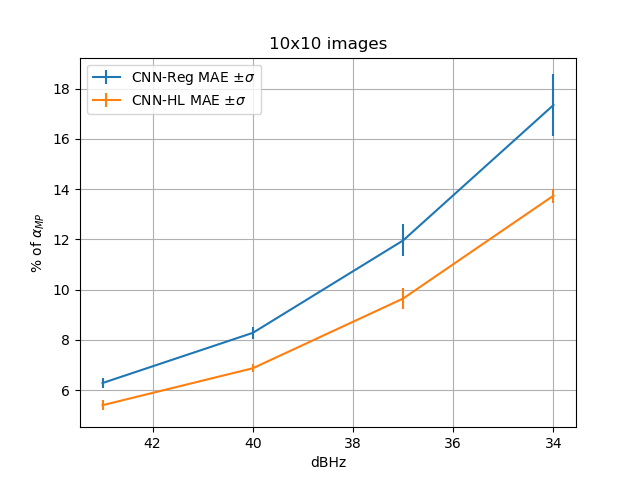} \\
    \end{tabular}
    \centering
    \caption{MAE behaviour according to \CNZERO ratio for $\alpha_{\mathit{MP}}$ estimation.}
	\label{fig:fig_alpha}
\end{figure}

\begin{figure}
    \centering
    \begin{tabular}{cc}
    \includegraphics[width=\mylength]{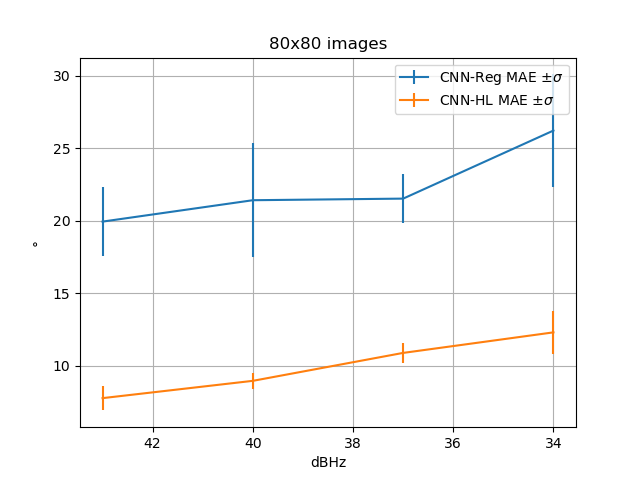} & \includegraphics[width=\mylength]{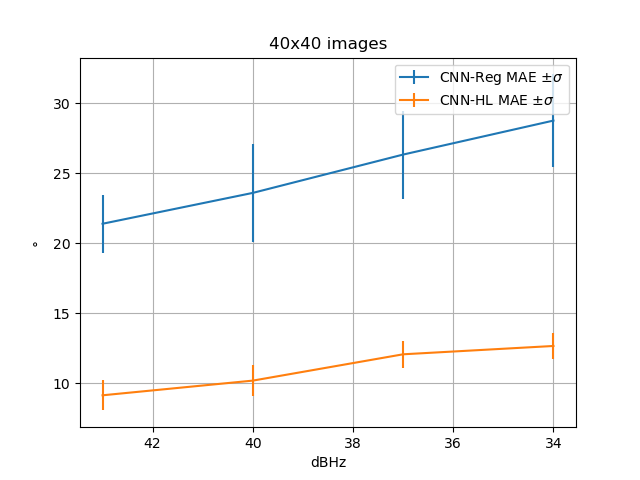} \\
    \includegraphics[width=\mylength]{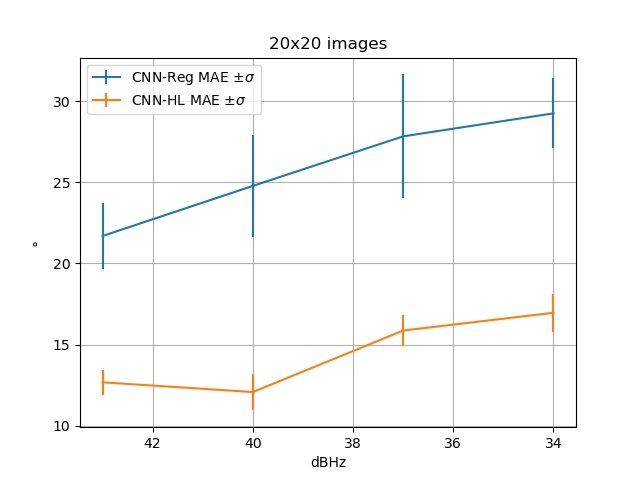} &
    \includegraphics[width=\mylength]{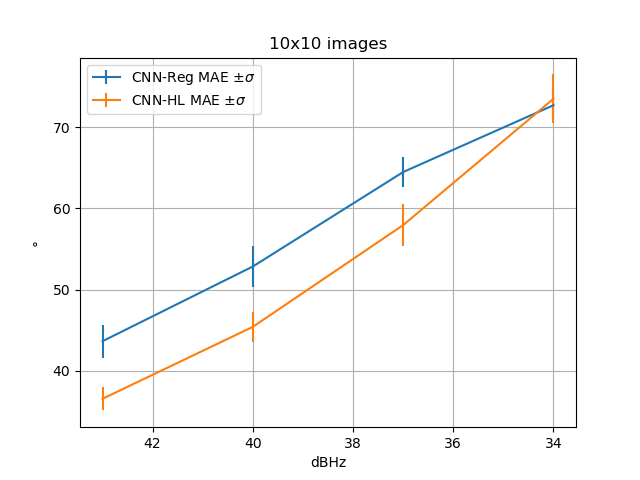} \\
    \end{tabular}
    \centering
    \caption{\gls{MAE} behaviour according to \CNZERO ratio for $\phi_{\mathit{MP}}$ estimation.}
	\label{fig:fig_phi}
\end{figure}

Figures~\ref{fig:fig_delay}, \ref{fig:fig_dopp}, \ref{fig:fig_alpha} and \ref{fig:fig_phi}, established with the second group of datasets, illustrate the \gls{CNN-Reg} and \gls{CNN-Reg} \gls{MAE} behaviour when \CNZERO varies. The curves are plotted according to a decreasing \CNZERO because a high \CNZERO corresponds to better receiving conditions. The scale on the x-axis starts from 43 \dBHz (good receiving conditions) and goes down to 34 \dBHz (deteriorated receiving conditions).

\subsection{Results discussion}
In Table~\ref{tab:delay_table}, it can be observed that in all conditions and datasets, \gls{CNN-HL} always shows better results than \gls{CNN-Reg}. The delay $\tau_{\mathit{MP}}$ estimation performance is higher with \gls{CNN-HL} on 20x20 images than the performance of \gls{CNN-Reg} on 80x80 images. Additionally, as expected the average \gls{MAE} and its standard deviation increase as the image resolution diminishes. 
In Table~\ref{tab:doppler_table}, similar results are observed for strong multi-paths. For moderate and weak multi-paths, the Doppler frequency estimation performance remains better on 40x40 images with \gls{CNN-HL} than the performance of \gls{CNN-Reg} on 80x80 images.
In Tables~\ref{tab:alpha_table} and \ref{tab:phase_table}, similar behaviour as $\tau_{\mathit{MP}}$ estimation are found. It can be noticed that in the particular case of $\alpha_{\mathit{MP}}$ estimation, the average \gls{MAE} decreases as the strength of the multi-path decreases. This could be explained by the fact that in strong, moderate and weak multi-path datasets, the $\alpha_{\mathit{MP}}$ parameter estimation is biased by the dataset design that only contains ranges of $\alpha_{\mathit{MP}}$ parameters.
In Table~\ref{tab:phase_table}, $\phi_{\mathit{MP}}$ estimation with both \gls{CNN} shows difficulties. For weak multi-paths on 40x40 images, the average \gls{MAE} remains at a level of $14^{\circ}$ which might not be sufficient for practical multi-path mitigation use.

\begin{small}
    \begin{threeparttable}
    	\caption{Average \gls{MAE} $\tau_{\mathit{MP}}$ estimation error in $10^{-2}$ \Tc for different multi-path attenuations.}
	    \label{tab:delay_table}
	    \begin{tabular*}{\linewidth}{@{\extracolsep{\fill}}llll}
		    \toprule
		    Dataset & \multirow{2}{1cm}{Image size} & \multicolumn{2}{c}{Algorithm}\\
		    & & \gls{CNN-Reg} & \gls{CNN-HL} \\
		    \midrule
		    \multirow{4}{2cm}{Strong multi-paths}      
		    & 80x80 & 2.50 $\pm$ 0.22 & \textbf{0.64} $\pm$ 0.18\\
		    & 40x40 & 2.91 $\pm$ 0.32 & \textbf{0.76} $\pm$ 0.11\\
		    & 20x20 & 4.30 $\pm$ 0.42 & \textbf{1.28} $\pm$ 0.20\\
		    & 10x10 & 6.70 $\pm$ 0.40 & \textbf{4.85} $\pm$ 0.17\\\hline
		    \multirow{4}{2cm}{Moderate multi-paths}      
		    & 80x80 & 3.03 $\pm$ 0.54 & \textbf{0.66} $\pm$ 0.04\\
		    & 40x40 & 3.44 $\pm$ 0.65 & \textbf{0.96} $\pm$ 0.26\\
		    & 20x20 & 4.44 $\pm$ 0.45 & \textbf{1.63} $\pm$ 0.15\\
		    & 10x10 & 9.90 $\pm$ 0.72 & \textbf{7.35} $\pm$ 0.38\\\hline
		    \multirow{4}{2cm}{Weak multi-paths}      
		    & 80x80 & 4.64 $\pm$ 1.44 & \textbf{1.01} $\pm$ 0.13\\
		    & 40x40 & 4.33 $\pm$ 0.42 & \textbf{1.55} $\pm$ 0.20\\
		    & 20x20 & 7.20 $\pm$ 1.55 & \textbf{2.86} $\pm$ 0.40\\
		    & 10x10 & 26.13 $\pm$ 1.96 & \textbf{21.17} $\pm$ 0.58\\\hline
		    \bottomrule
	    \end{tabular*}
    \end{threeparttable}
\end{small}

\begin{small}
    \begin{threeparttable}
	    \caption{Average \gls{MAE} $f_{\mathit{MP}}$ estimation error in \Hz for different multi-path attenuations.}
	    \label{tab:doppler_table}
	    \begin{tabular*}{\linewidth}{@{\extracolsep{\fill}}llll}
		    \toprule
		    Dataset & \multirow{2}{1cm}{Image size} & \multicolumn{2}{c}{Algorithm}\\
    		& & \gls{CNN-Reg} & \gls{CNN-HL}\\
	        \midrule
		    \multirow{4}{2cm}{Strong multi-paths}      
		    & 80x80 & 1.22 $\pm$ 0.20 & \textbf{0.60} $\pm$ 0.04\\
		    & 40x40 & 1.60 $\pm$ 0.27 & \textbf{0.75} $\pm$ 0.07\\
		    & 20x20 & 2.36 $\pm$ 0.41 & \textbf{1.17} $\pm$ 0.12\\
		    & 10x10 & 5.16 $\pm$ 0.26 & \textbf{3.60} $\pm$ 0.18\\\hline
		    \multirow{4}{2cm}{Moderate multi-paths}      
		    & 80x80 & 1.33 $\pm$ 0.29 & \textbf{0.72} $\pm$ 0.06\\
		    & 40x40 & 1.70 $\pm$ 0.23 & \textbf{0.87} $\pm$ 0.06\\
		    & 20x20 & 2.61 $\pm$ 0.45 & \textbf{1.52} $\pm$ 0.11\\
		    & 10x10 & 7.08 $\pm$ 0.36 & \textbf{5.52} $\pm$ 0.27\\\hline
		    \multirow{4}{2cm}{Weak multi-paths}      
		    & 80x80 & 1.75 $\pm$ 0.31 & \textbf{1.26} $\pm$ 0.15\\
		    & 40x40 & 2.41 $\pm$ 0.38 & \textbf{1.42} $\pm$ 0.13\\
		    & 20x20 & 4.30 $\pm$ 0.41 & \textbf{2.84} $\pm$ 0.45\\
		    & 10x10 & 21.48 $\pm$ .68 & \textbf{19.04} $\pm$ 1.07\\\hline
		    \bottomrule
	    \end{tabular*}
    \end{threeparttable}
\end{small}

\begin{figure}
    \centering
    \includegraphics[width=\linewidth]{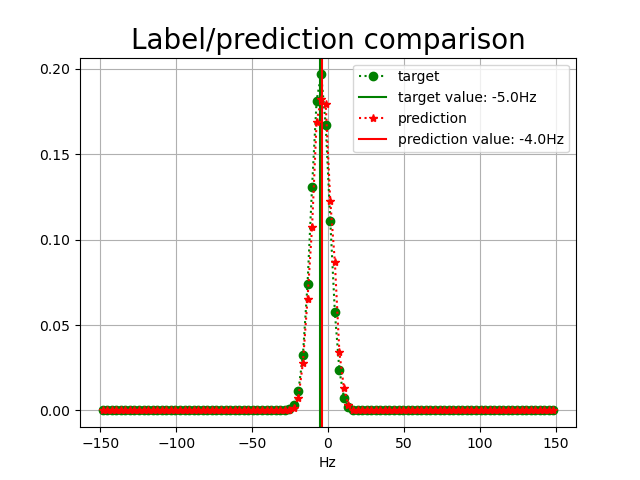}
    \caption{Comparison between the target and the prediction distribution probabilities for a correct prediction.}
	\label{fig:fig_correct_prediction}
\end{figure}

\begin{figure}
    \centering
    \includegraphics[width=\linewidth]{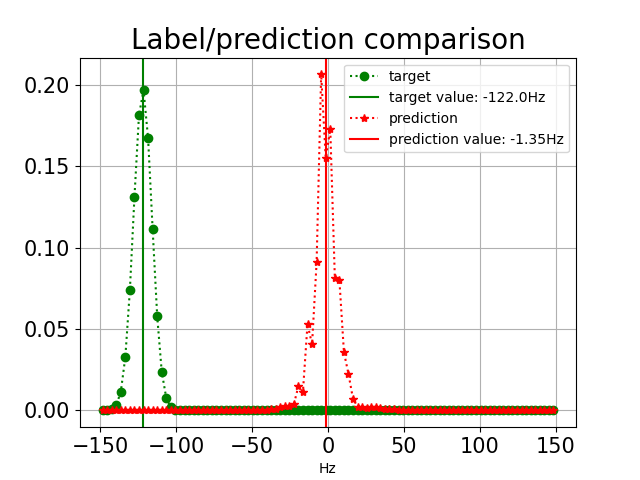}
    \caption{Comparison between the target and the prediction distribution probabilities for a wrong prediction.}
	\label{fig:fig_wrong_prediction}
\end{figure}

In Figures~\ref{fig:fig_correct_prediction} and \ref{fig:fig_wrong_prediction}, a comparison of label and prediction distribution probabilities are shown. They correspond to a correct and a wrong estimation respectively. Each point of the plots represents the output of one of the $K$ neurons from the \gls{CNN-HL} softmax layer. It can be noticed that in Figure~\ref{fig:fig_correct_prediction}, both curves are nearly superposed and both form a near Gaussian distribution. In the opposite, in Figure~\ref{fig:fig_wrong_prediction}, the target and predicted probability distribution do not coincide. This situation corresponds to a wrong estimation. In this case, the Gaussian behaviour is not so well reconstructed.

 \begin{small}
    \begin{threeparttable}
	    \caption{$\alpha_{\mathit{MP}}$ estimation error in \% for different multi-path attenuations.}
	    \label{tab:alpha_table}
	    \begin{tabular*}{\linewidth}{@{\extracolsep{\fill}}llll}
		    \toprule
		    Dataset & \multirow{2}{1cm}{Image size} & \multicolumn{2}{c}{Algorithm}\\
    		& & \gls{CNN-Reg} & \gls{CNN-HL}\\
	    	\midrule
		    \multirow{4}{2cm}{Strong multi-paths}      
		    & 80x80 & 3.95 $\pm$ 0.93 & \textbf{1.30} $\pm$ 0.26\\
		    & 40x40 & 4.45 $\pm$ 0.61 & \textbf{1.48} $\pm$ 0.14\\
		    & 20x20 & 5.60 $\pm$ 0.70 & \textbf{2.22} $\pm$ 0.20\\
		    & 10x10 & 7.32 $\pm$ 0.40 & \textbf{5.99} $\pm$ 0.25\\\hline
		    \multirow{4}{2cm}{Moderate multi-paths}      
		    & 80x80 & 3.71 $\pm$ 0.64 & \textbf{1.08} $\pm$ 0.05\\
		    & 40x40 & 3.76 $\pm$ 0.53 & \textbf{1.22} $\pm$ 0.15\\
		    & 20x20 & 4.79 $\pm$ 0.55 & \textbf{1.73} $\pm$ 0.10\\
		    & 10x10 & 5.97 $\pm$ 0.16 & \textbf{5.00} $\pm$ 0.07\\\hline
		    \multirow{4}{2cm}{Weak multi-paths}      
		    & 80x80 & 3.53 $\pm$ 0.88 & \textbf{1.91} $\pm$ 1.16\\
		    & 40x40 & 4.06 $\pm$ 1.28 & \textbf{1.38} $\pm$ 0.57\\
		    & 20x20 & 4.09 $\pm$ 1.10 & \textbf{1.66} $\pm$ 0.22\\
		    & 10x10 & 7.87 $\pm$ 0.33 & \textbf{6.08} $\pm$ 0.12\\\hline
		    \bottomrule
	    \end{tabular*}
    \end{threeparttable}
\end{small}

\begin{small}
    \begin{threeparttable}
	    \caption{$\phi_{\mathit{MP}}$ estimation error in degrees for different multi-path attenuations.}
	    \label{tab:phase_table}
	    \begin{tabular*}{\linewidth}{@{\extracolsep{\fill}}llll}
		    \toprule
		    Dataset & \multirow{2}{1cm}{Image size} & \multicolumn{2}{c}{Algorithm}\\
		    & & \gls{CNN-Reg} & \gls{CNN-HL}\\
		    \midrule
		    \multirow{4}{2cm}{Strong multi-paths}      
		    & 80x80 & 12.30 $\pm$ 1.46 & \textbf{4.19} $\pm$ 0.81\\
		    & 40x40 & 15.41 $\pm$ 3.46 & \textbf{4.80} $\pm$ 0.67\\
		    & 20x20 & 15.59 $\pm$ 2.44 & \textbf{6.34} $\pm$ 0.69\\
		    & 10x10 & 29.68 $\pm$ 2.06 & \textbf{24.12} $\pm$ 1.11\\\hline
		    \multirow{4}{2cm}{Moderate multi-paths}      
		    & 80x80 & 15.71 $\pm$ 3.07 & \textbf{5.51} $\pm$ 0.89\\
		    & 40x40 & 15.72 $\pm$ 1.74 & \textbf{5.71} $\pm$ 0.88\\
		    & 20x20 & 17.90 $\pm$ 1.99 & \textbf{8.46} $\pm$ 1.03\\
		    & 10x10 & 42.55 $\pm$ 2.30 & \textbf{34.17} $\pm$ 2.02\\\hline
		    \multirow{4}{2cm}{Weak multi-paths}      
		    & 80x80 & 26.45 $\pm$ 6.25 & \textbf{10.23} $\pm$ 0.84\\
		    & 40x40 & 30.06 $\pm$ 3.35 & \textbf{14.12} $\pm$ 1.28\\
		    & 20x20 & 29.57 $\pm$ 3.17 & \textbf{17.15} $\pm$ 2.01\\
		    & 10x10 & 70.23 $\pm$ 1.56 & \textbf{68.82} $\pm$ 2.27\\\hline
		    \bottomrule
	    \end{tabular*}
    \end{threeparttable}
\end{small}

\subsection{Results synthesis}
The experiments made on the various datasets show the boosting effect of the \gls{CNN-HL} algorithm which always performed better than \gls{CNN-Reg} with not only a lower average \gls{MAE} but also a lower standard deviation. The gain brought by the distributional technique allows more accurate and precise estimation on smaller images. The input size could be reduced from a resolution of 80x80 to 40x40 and sometimes 20x20. The results tend to indicate that the multi-path phase estimation is a more difficult task and the level of performance achieved by the models might not be sufficient in practice.

\section{Conclusions}\label{conclusions}
Estimating information from images is a useful but difficult task. In this work, we have addressed deep learning regression from multiple images channels. The proposed model makes use of convolutional neural layers in order to extract high level features from images. Instead of carrying out classical regression on these extracted features, a soft labelling approach is used to learn underlying target distributions. The idea is to allow the neural network model to account for some uncertainty in the target and therefore increase its generalization performance. The target distributions are modeled using histograms and a specific histogram loss function based on the \gls{KL} divergence is applied during training. The resulting neural architecture incorporates a softmax output in order to reconstruct the histogram discrete target probability. The complete process could be applied to any applications that requires inference of function values from multiple images channels. The model is applied to \gls{GNSS} multi-path estimation where multi-path signal parameters have to be estimated from correlator output images from the I and Q channels. The multi-path signal delay, attenuation, Doppler frequency and phase parameters are estimated from synthetically generated datasets of satellite signals. Experiments are conducted under various receiving conditions and various input images resolutions. For all receiving conditions that have been tested, the proposed soft labelling \gls{CNN} technique using distributional loss outperforms classical \gls{CNN} regression. In addition, the gain in accuracy obtained by the model allows downsizing of input image resolution from 80x80 down to 40x40 or sometimes 20x20. This reduction of image resolution is a first step towards the implementation of such models in physical receivers that are limited in the number of correlator outputs that can be designed. Despite real datasets are difficult to construct and label, further research using the \gls{CNN-HL} model on \gls{GNSS} data should focus on data that incorporates real receiving conditions. Additionally, from a model perspective, future investigation should propose adaptive histogram loss techniques that adjust the number of bins and target distribution parameters to the data.

\section*{Acknowledgements}
This project has been funded by Oktal-Synthetic Environment under the ANRT PhD program.

\bibliographystyle{IEEEtran}
\bibliography{refs}

\end{document}